\documentclass[letterpaper, 10 pt, conference]{ieeeconf}  

\IEEEoverridecommandlockouts                              
\usepackage{amsmath}
\usepackage{amsfonts}
\usepackage{amssymb}
\usepackage{color}

\usepackage{graphics} 
\usepackage{epsfig} 
\usepackage{mathptmx} 
\usepackage{times} 
\usepackage{amsmath} 
\usepackage{amssymb}  

\title{\LARGE \bf
Bayesian Scale Estimation for Monocular SLAM Based on Generic Object Detection for Correcting Scale Drift
}

\author{Edgar Sucar$^{1}$ and Jean-Bernard Hayet$^{1}$
\thanks{$^{1}$Centro de Investigaci\'on en Matem\'aticas,
        Guanajuato, M\'exico
        {\tt\small \{edgar.sucar,jbhayet\}@cimat.mx}}
}

\begin{document}

\maketitle
\thispagestyle{empty}
\pagestyle{empty}

\begin{abstract}
This work proposes a new, online algorithm for estimating the local scale correction to apply to the output of a monocular SLAM system and obtain an as faithful as possible metric reconstruction of the 3D map and of the camera trajectory. Within a Bayesian framework, it integrates observations from a deep-learning based generic object detector and a prior on the evolution of the scale drift. For each observation class, a predefined prior on the heights of the class objects is used. This allows to define the observations likelihood. Due to the scale drift inherent to monocular SLAM systems, we integrate a rough model on the dynamics of scale drift. Quantitative evaluations of the system are presented on the KITTI dataset, and compared with different approaches. The results show a superior performance of our proposal in terms of relative translational error when compared to other monocular systems.
\end{abstract}

\section{Introduction}

Monocular simultaneous localization and mapping (SLAM) is a classical problem that has been tackled in various forms in the robotics and computer vision communities for more than 15 years. Starting from the seminal work of Davison~\cite{Davison2003}, impressive results have been obtained in the construction of sparse or semi-dense 3D maps and in visual odometry~\cite{Forster14,murTRO2015,Engel14}, with  a single camera. Given the availability and low price of this kind of sensor, many applications have been developed on top of monocular SLAM systems.  

\begin{figure}[t!]
	\begin{center}
    \includegraphics[width=0.6\linewidth]{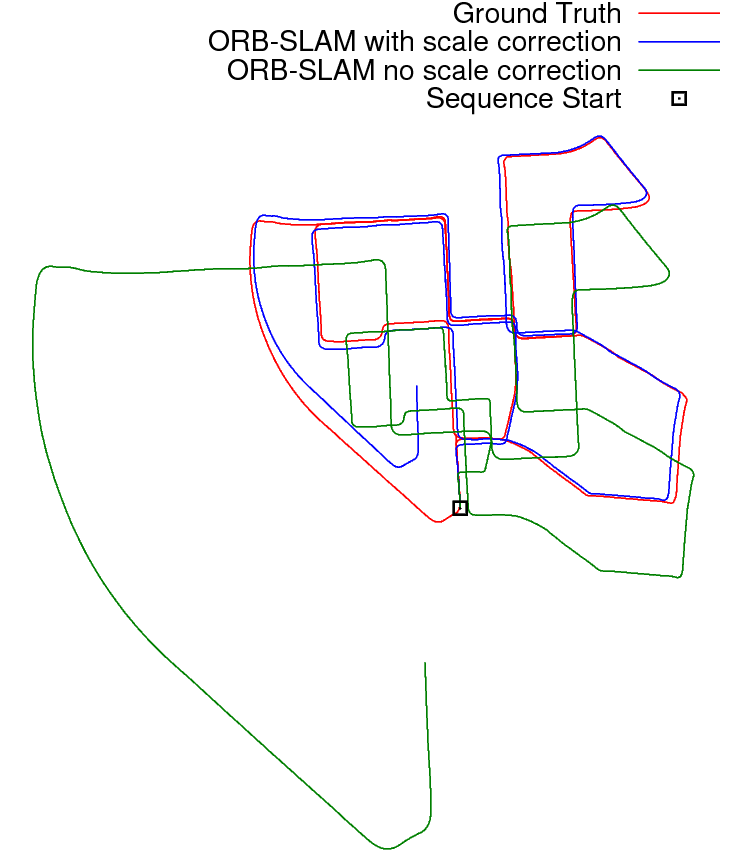}
     \vspace{1em}
	\includegraphics[width=0.9\linewidth]{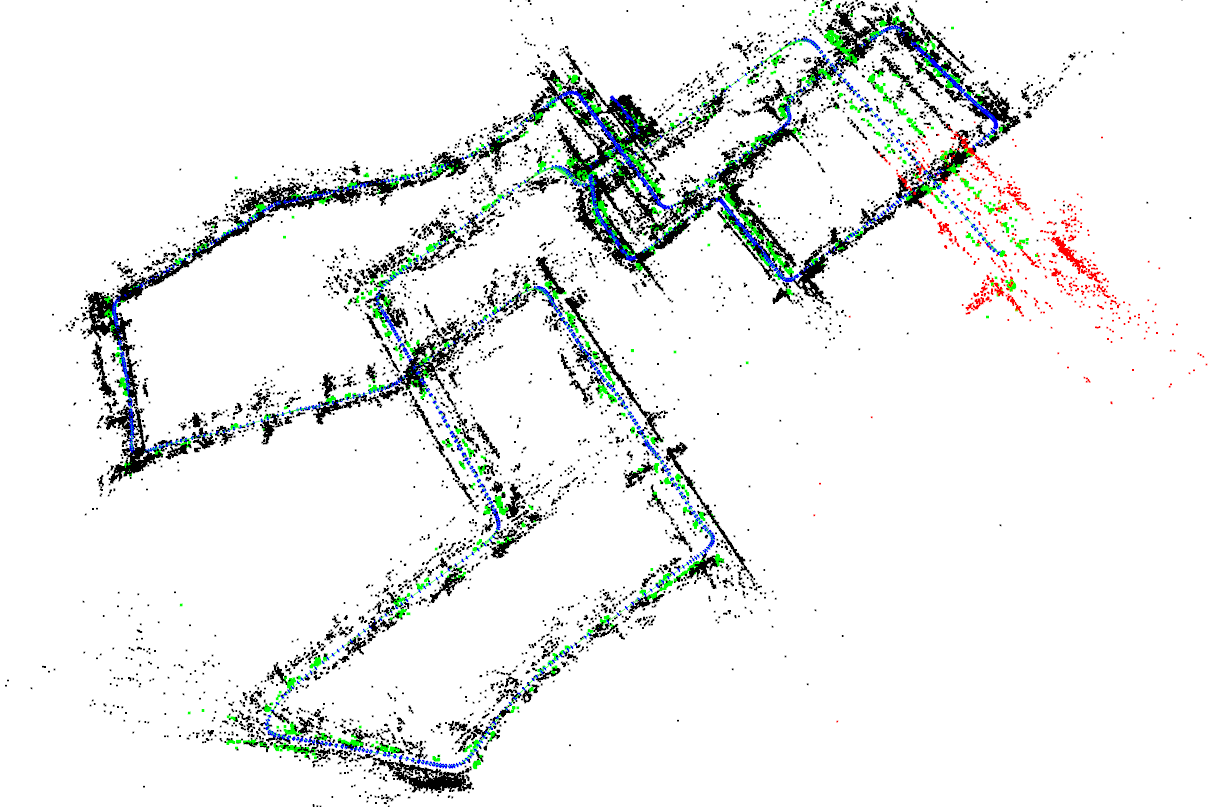}
    \vspace{1em}
	\includegraphics[width=0.9\linewidth]{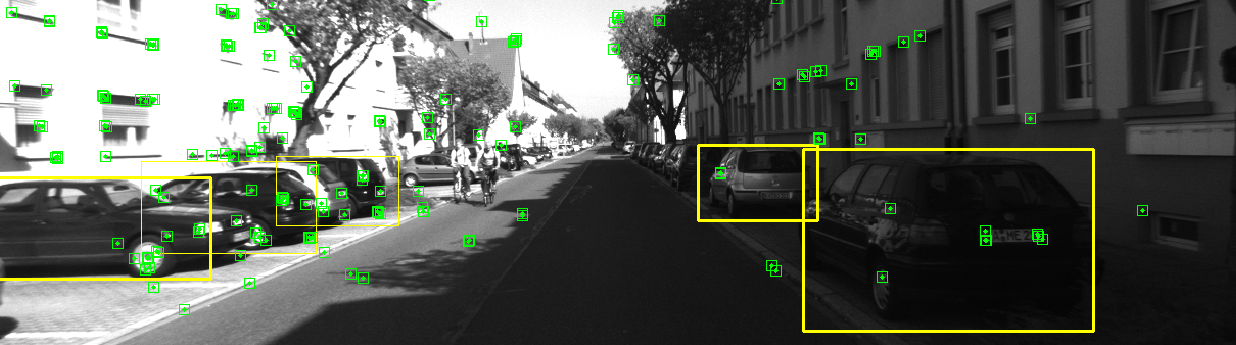}
		\end{center}
\caption{We estimate the scale correction to apply to the camera trajectory and the map (top and middle figures), using Bayesian inference based on a detector of instances of pre-defined object classes (e.g. cars, bottom figure), with prior distributions specified on the height of theses objects. The top figure shows, for KITTI sequence $00$, the reconstructed trajectory (without loop closure) by ORB-SLAM, in green, while our corrected trajectory is depicted in blue, and the ground truth in red.\label{fig:summary}}
\end{figure}

One of the main limits of monocular SLAM systems is that, because of the projective nature of the sensor, the scale of the 3D scene is not observable. This has two important implications: (1) The scale of the camera trajectory and of the reconstructed map are arbitrary, depending typically on choices made during the system initialization; (2) While no loop closure process is applied on the map and on the trajectory (usually with some form of bundle adjustment), the scale error may drift without bound. For example, in Fig.~\ref{fig:summary}, top, the basic version of ORB-SLAM (without loop closure) outputs the green path on one of the KITTI dataset urban video sequences. The ground truth appears in red. The scale drift explains why the internal scale estimate is clearly increasing during the whole experiment. When loop closure processes are applied, the global scale is made coherent over the map and the trajectory, but again, at an arbitrary value. Since for many applications (mobile robotics, augmented reality,\dots) the true scale factor plays a critical role, automatic methods to infer it are important. 

The main idea of this work is that, based on the semantic content of what a monocular system can perceive, and even if each perceived cue gives uncertain evidence on scale, a robot should be able to infer the global scale of the structures present in the scene. Handling the potential contradictions between cues can be done efficiently within a Bayesian framework, as it allows to specify and fuse nicely the uncertain knowledge given by each visual cue. Evidence of this inference process in animal visual systems has been exhibited~\cite{perception}. As an example, when it observes a scene containing cars and houses, the human brain, based on its prior knowledge on the size of typical cars and houses, can infer depths and distances,  even though there is a slight possibility that all of these objects are just small objects in a toy world. Based on this idea of using general semantic knowledge (e.g., detections of cars at the bottom of Fig.~\ref{fig:summary}), we build up a Bayesian inference system on the monocular SLAM scale correction. This system allows us to produce, in the case of Fig.~\ref{fig:summary}, the blue path, much closer to the ground truth than the green one (without correction).   

We review related work in Section~\ref{sec:related}, then give an overall explanation of our algorithm in Section~\ref{sec:overall}, and give specific details on the probability distributions that we use in Section~\ref{sec:distribs}. Quantitative and qualitative evaluations are given in Section~\ref{sec:experiments}.

\section{Related work}
\label{sec:related}
Monocular SLAM has been a tool of choice in 3D scene and camera trajectory reconstruction, e.g. in mobile robotics or augmented reality, in particular because monocular systems are widespread and inexpensive. Two categories of online techniques coexist in the literature for this purpose: the ones that use Bayesian filtering~\cite{Davison2003} and the ones that extend the traditional bundle adjustment algorithms to online systems~\cite{Klein2007,murORB2}. The latter have allowed to attain outstanding results in the recent years, at much larger scales than the former. However, common limitations of all the existing methods are that: (i) the scale of the reconstruction is unknown by essence, and (ii) the consecutive reconstructions/pose estimations may introduce scale drift which makes the global maps or the complete trajectories inconsistent. Most of the classical SLAM systems~\cite{Davison2003,murTRO2015} use ad-hoc elements in the initialization phase to set the reconstruction scale at the beginning, i.e., known objects or known motions. To limit the scale drift, loop closure techniques allow to reset the scale in a consistent way with its initial value~\cite{murTRO2015}. 

An obvious solution to the scale recovery problem is to upgrade the sensors to devices capable of measuring depth (e.g., Kinect~\cite{kinectfusion}) or displacements (e.g., IMU sensors~\cite{IMU}), but this may be costly or simply unfeasible. In this work, we focus on  using only the semantic content of RGB images, together with prior uncertain information on this content, to infer the global scale of the reconstruction. Previous works in this direction include~\cite{GalvezLopez2016}, where the output of an object recognition system is used in the map/trajectory optimization, and~\cite{Salas2013}, where object detection was used to simplify the map building process with depth cameras. In both cases, databases of specific instances of objects were used, whereas our work uses more general object classes.

Closer to our approach,~\cite{ismarscale} proposes a scale estimation method that tracks the user face and uses it as a cue for determining the scale. The method is designed for cell phones equipped with front and rear cameras and would be difficult to extend to more generic monocular systems. In~\cite{Song13}, ~\cite{Zhou2016ReliableSE}, and ~\cite{7225730} in a context of monocular vision embedded on cars, the scale is integrated based on the knowledge of the camera height above the ground, and based on local planarity assumptions in the observed scene. Again, our method can be applied in more generic settings, although we evaluate here in this road navigation context.  

In~\cite{Frost2016}, the approach is similar to ours as it also uses size priors for the detections and as it is applied to urban scenes. Nevertheless, we do not rely on consecutive object detections (which implies data association to be solved) and instead get observations from any object detection on which projections of the reconstructed 3D cloud points lie. Additionally, by relying only on points projected from the 3D map, we ensure in some way not to include information coming from dynamic objects. We use a Bayesian formulation that allows us to integrate different elements of previous knowledge, such as a prior on the variations of the scale correction factor.    

Finally, our work is also reminiscent of approaches that perform machine learning-based depth inference from the texture of monocular images~\cite{Saxena2009}. Here we combine the strength of recent deep learning detection techniques~\cite{yolo} with the power and flexibility of Bayesian inference, so as to integrate available prior knowledge in a principled way.  

In a first version of this work~\cite{Sucar2017}, we adopted a similar strategy to the one presented here, but this paper introduces three novel contributions:
\begin{itemize}
\item the estimated scale correction parameter is now associated to a motion model, and its variations are related to the SLAM system scale drift (see Section~\ref{subsection:trans}),
\item a new, more robust, probabilistic observation model is proposed (see Section~\ref{subsection:likelihood}).
\item it is now implemented in a state of the art Monocular SLAM system, which allows for improved evaluation.
\end{itemize}

\section{Detection-based scale estimation}
\label{sec:overall}
In this section, we present the core elements of our detection-based scale correction system.

\subsection{Notations and definitions}

From now on, we assume that we run a monocular SLAM algorithm (such as~\cite{murTRO2015}). We denote the camera calibration matrix by $\textbf{K}$. The camera pose at time $k$ is referred to as $\mathbf{T}_k \in SE(3)$. $3D$ points, reconstructed by the SLAM algorithm, are indexed by $j$ and referred to as $p_w^j$ in the world frame and as $p_c^j = \mathbf{T}_k p_w^j$ in the camera frame. Points projected on the image frame at time $k$ are noted as $\pi_k^j = \pi(\textbf{K},p_c^j)$ where $\pi$ is the standard perspective projection function.

As explained hereafter, we rely on a generic object detector that, given an image in our video sequence, outputs a list of detected objects, together with the class they belong to. This detector (see Section~\ref{subsection:descriptionexp}) has been trained to detect instances from dozens of classes. In frame $k$, we denote the set of object detections as $\mathcal{D}_k$, and the set of sets of detections done at frames $1,..,k$ as $\mathcal{D}_{1:k}$. Each individual detection is noted as $D_k^i \in \mathcal{D}_k$. We define two functions $R()$ and $c()$, such that $R(D_k^i)$ is the rectangular region in the image corresponding to the detection, like the rectangles depicted in Fig.~\ref{fig:summary}, and $c(D_k^i)$ is the object class of the detection (e.g., ``truck'', ``car'', ``bottle''\dots). Finally, we introduce a prior height distribution built beforehand for an object class $c(D_k^i)$ as $p_{c(D_k^i)}(H)$ defined on $\mathbb{R}^{+}$.

 A system such as ORB-SLAM~\cite{murTRO2015} maintains a local map $\mathcal{N}_k$ of points expressed in the world frame. It is a subset of the global map that contains the set of points from keyframes $\mathcal{K}_1$ that share map points with the current frame, and the set $\mathcal{K}_2$ with neighbors in $\mathcal{K}_1$ in the covisibility graph. The local map is used for tracking purposes and it is optimized via bundle adjustment every time a new keyframe is added. Most other SLAM systems work in a similar way. Our aim in this work is to estimate the scalar $\kappa_k$, which we define as the correction to apply to the local 3D map or to the local trajectory in order to obtain the correct scale of the local map maintained by a visual SLAM system, at time $k$.

\subsection{Problem statement}

We model $\kappa_k$ as a random variable to be estimated at time $k$ such that for any pair of points $p_w^i, p_w^j \in \mathcal{N}_k$, the true metric distance $D_r$ between them is given by  $D_r(p_w^i, p_w^j) = \kappa_k D(p_w^i, p_w^j) + \nu$, where $D$ is the distance measure in the current reconstruction and $\nu$ is a reconstruction error noise.

Since the local map is used for tracking, we can recover the camera trajectory with its correct metric scale by estimating $\kappa_k$. Given $\mathbf T_k$, the pose of the camera at time $k$ according to the visual SLAM system, the pose $\mathbf{\tilde{T}}_k$ with its correct metric scale can then be computed incrementally by 
$$
\mathbf{\tilde{T}}_k = s(\mathbf{T}_k \mathbf{T}^{-1}_{k-1},\kappa_k) \mathbf{\tilde{T}}_{k-1}
$$

\noindent
where $s(\mathbf T,\alpha)$ builds a similarity from the rigid transformation $\mathbf T$ and the scale factor $\alpha$.

In the following, we develop a Bayesian formulation for the estimation of this local scale correction as the mode of the posterior distribution:

$$
p(\kappa_k|\mathcal{N}_{1:k},\mathcal{D}_{1:k}),
$$

\noindent
i.e., conditioned to the observation of detected objects $\mathcal{D}_{1:k}$ and to the SLAM local reconstructions $\mathcal{N}_{1:k}$.    
\subsection{Bayesian framework for estimating the scale correction}

As mentioned above, we stress that, because of the scale drift inherent to monocular SLAM systems, the global scale correction $\kappa$ is varying with time. To estimate it, we use observations from object detections on which we have priors for their belonging classes (e.g., priors on cars heights in Fig.~\ref{fig:summary}), and we use a dynamical model to cope with potential variations in the internal scale of the SLAM algorithm, i.e., a rough model on the dynamics of scale drift. As we do not have a detailed knowledge of these variations (it probably depends on the internal logic of each SLAM algorithm), we use a simple dynamic model from frame $k$ to frame $k+1$

\begin{equation}
\kappa_{k+1} = \kappa_{k} +\nu
\label{eq:transmodel}
\end{equation}

\noindent
where $\nu\sim \mathcal{N}(0,\sigma^p_k)$. We will explain how to select $\sigma^p_k$ in Section~\ref{subsection:trans}.

In frame $k$, let us suppose that we are capable of getting a set of object detections $\mathcal{D}^k$. By applying  Bayes formula,  
	\begin{small}
		\begin{equation}
		\begin{split}
p(\kappa_k|\mathcal{N}_{1:k},\mathcal{D}_{1:k}) \propto p(\mathcal{D}^k|\kappa_k,\mathcal{N}_{1:k},\mathcal{D}_{1:k-1})p(\kappa_k|\mathcal{N}_{1:k},\mathcal{D}_{1:k-1}).
        \end{split}
		\end{equation}
	\end{small}    
    
For the sake of clarity in this derivation, let us first suppose that $\mathcal{D}_k$ consists of a single detection $D_k^1$ of an object belonging to class $c(D_k^1)$. 

$$
\begin{array}{l}      p(\kappa_k|\mathcal{N}_{1:k},\mathcal{D}_{1:k})
\\
\propto p(\kappa_k|\mathcal{N}_{1:k},\mathcal{D}_{1:k-1})\int_{H} p(H,D_k^1|\kappa_k,\mathcal{N}_{1:k},\mathcal{D}_{1:k-1}) dH \\
\propto  p(\kappa_k|\mathcal{N}_{1:k},\mathcal{D}_{1:k-1}) \int_{H} p(D_1|H,\kappa_k,\mathcal{N}_{1:k},\mathcal{D}_{1:k-1}) \\ p(H|\kappa_k,\mathcal{N},\mathcal{D}^{1:k-1})dH  \\
	\propto   p(\kappa_k|\mathcal{N}_{1:k},\mathcal{D}_{1:k-1}) \int_{H} p(D_k^1|H,\kappa_k,\mathcal{N}_{1:k}) p_{c(D_k^1)}(H)dH,\\
\propto  \int_{\kappa_{k-1}} p(\kappa_k|\kappa_{k-1}) p(\kappa_{k-1}|\mathcal{N}_{1:k-1},\mathcal{D}_{1:k-1}) \\
\int_{H} p(D_k^1|H,\kappa_k,\mathcal{N}_{k}) p_{c(D_k^1)}(H)dH. 
		\end{array}
$$

Through the formula above, we obtain a recursive Bayes filter that allows to make updates of the scale correction estimate at each new frame, by incorporating three terms: (1) a transition probability $p(\kappa_k|\kappa_{k-1})$ that models the scale drift in the SLAM algorithm; (2) a likelihood term $p(D_k^1|H,\kappa_k,\mathcal{N}_{k})$ that evaluates the probability of having the observed detection, given a current point cloud $\mathcal{N}_k$ built by the SLAM algorithm, given a possible height $H$ for the detected object of class $c(D_k^1)$, and given a global scale correction $\kappa_k$; (3) a prior on heights $p_{c(D_k^1)}(H)$, specific to the class of the detected object $c(D_k^1)$.

This means that, at each step, we can update the posterior on $\kappa_k$. We implemented the previous inference scheme in two ways: as a discrete Bayes filter and as a Kalman filter. Using one or the other depends mainly on the context and on the nature of the involved distributions. In the first case, we use a histogram representation for the posterior distribution and for $p_{c(D^1_k)}(H)$, the prior probability the object height. By discretizing the possible heights $H_{m}$ over a pre-defined interval, we can compute the likelihood term as $\sum_{m} p(D_k^1|H_{m},d,\mathcal{N}_k) p_{c_(D^1_k)}(H_{m})$. In the second case, when the involved distributions are Gaussian and the models linear, then we have an instance of the Kalman filter, which takes a simple form of mean/variance updates (see Section~\ref{subsection:Kalman}).

Note that in the more general case of $|\mathcal D_k|>1$, and by assuming conditional independence between the different detections observed in frame $k$, we have

\begin{equation}
\begin{split}	&p(\kappa_k|\mathcal{N}_k,\mathcal{D}^1,\dots,\mathcal{D}^k)\\ 
&\propto  \int_{\kappa_{k-1}} p(\kappa_k|\kappa_{k-1}) p(\kappa_{k-1}|\mathcal{N}_{1:k-1},\mathcal{D}_{1:k-1}) \\
&p(\kappa_k|\mathcal{N}_k,\mathcal{D}^1,\dots,\mathcal{D}^{k-1}) \prod_{l=1}^{D_k} \int_{H}  p(D_k^l|H,\kappa_k,\mathcal{N}_k) p_{c(D^l_k)}(H)dH.
\end{split}
\label{eq:update}
\end{equation}
In the following, we give details on these three distributions.

\section{Definition of the probabilistic models}
\label{sec:distribs}
\subsection{Transition probability}
\label{subsection:trans}
As stated before, the distribution $p(\kappa_k|\kappa_{k-1})$ allows us to encode time variations of the global scale correction. These variations are caused by accumulation of errors in the mapping and tracking threads of the SLAM algorithm.   

Experiments show that larger global scale variations occur in situations when the camera experiences greater angular displacement~\cite{murORB2}. For this reason, $p(\kappa_k|\kappa_{k-1})$ is modeled as a Gaussian distribution centered at $\kappa_{k-1}$ with a standard deviation $\sigma^p_k$, variable for each frame $k$, and proportional to the angular displacement of the camera.

Let $\omega_k$ be the angular displacement (in degrees) along the rotation axis between $\mathbf{T}_{k-1}$ and $\mathbf{T}_{k}$. Let $k_0$ be the last time since the scale was updated, then we define $\Omega_k = \sum_{i=k_0}^k \omega_i$. 

The standard deviation $\sigma^p_k$ is then calculated as 
\begin{equation}
\sigma^p_k = \sigma_{min} + \Omega_k \frac{\sigma_{max}}{\Omega_{max}}.
\label{eq::omega}
\end{equation}.

The values observed to work in practice are $\sigma_{min} = 0.00001$, $\sigma_{max} = 0.05$, and $\Omega_{max} = 120^{\circ}$. These values for $\sigma_{min}$ and $\sigma_{max}$ have been determined from the observed variations of the scale correction along several test sequences. 

\subsection{Likelihood of detections}
\label{subsection:likelihood}
The term $p(D_k^1|H,\kappa_k,\mathcal{N}_k)$ is the probability that the detected object has the dimensions in pixels with which it was detected, given that the object has a real size $H$, that the scale is $\kappa_k$, and that the local map is $\mathcal{N}_k$. 

The general idea to evaluate it is to estimate the height of the detected object using $R(D_k^1)$ and $\mathcal{N}_k$, then to obtain a scale correction estimate and compare it with $\kappa_k$.

Let $\{p_c^1,\dots, p_c^m \}$ be the points from the local map $\mathcal{N}_k$ transformed in the camera frame and whose projection lies inside $R(D_k^1)$, with the current pose and map parameters. We assume that in the world frame in which the SLAM system does its tracking and mapping, we can identify the vertical direction. We assume that the detected object surface is parallel to the vertical direction and that the object is oriented vertically. From $\{p_c^1, \dots, p_c^m \}$, we will first construct a point $p^s$ that will lie on a vertical straight line $\Lambda$ to be used to infer the object height. Let $\{\hat{p}_c^1, \dots, \hat{p}_c^m \}$ be the projection of the points $\{p_c^1, \dots, p_c^m \}$ in the plane perpendicular to the vertical direction and that pass through the camera position. 

\begin{figure*}[h!]
	\begin{center}
	\includegraphics[width=0.4\linewidth]{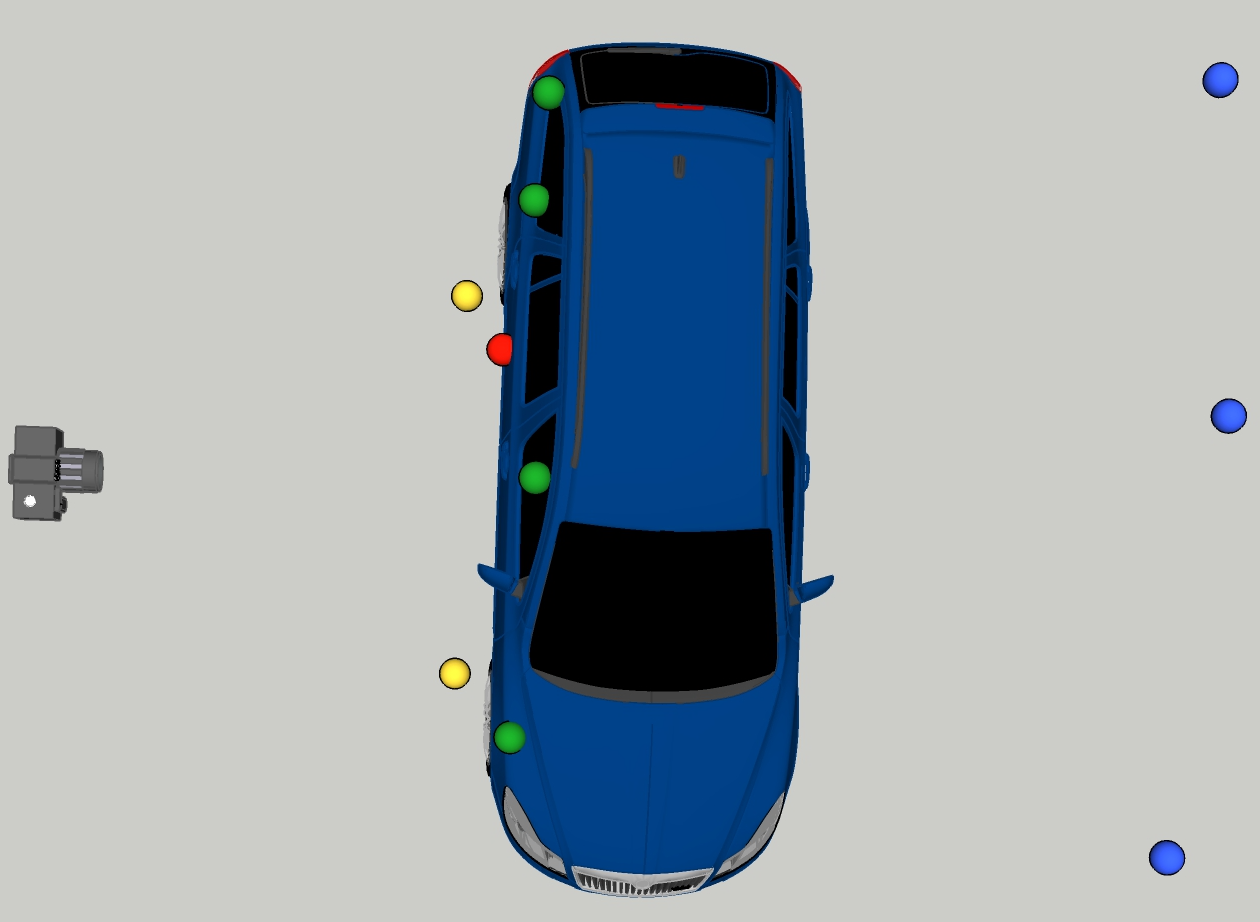}
	\includegraphics[width=0.55\linewidth]{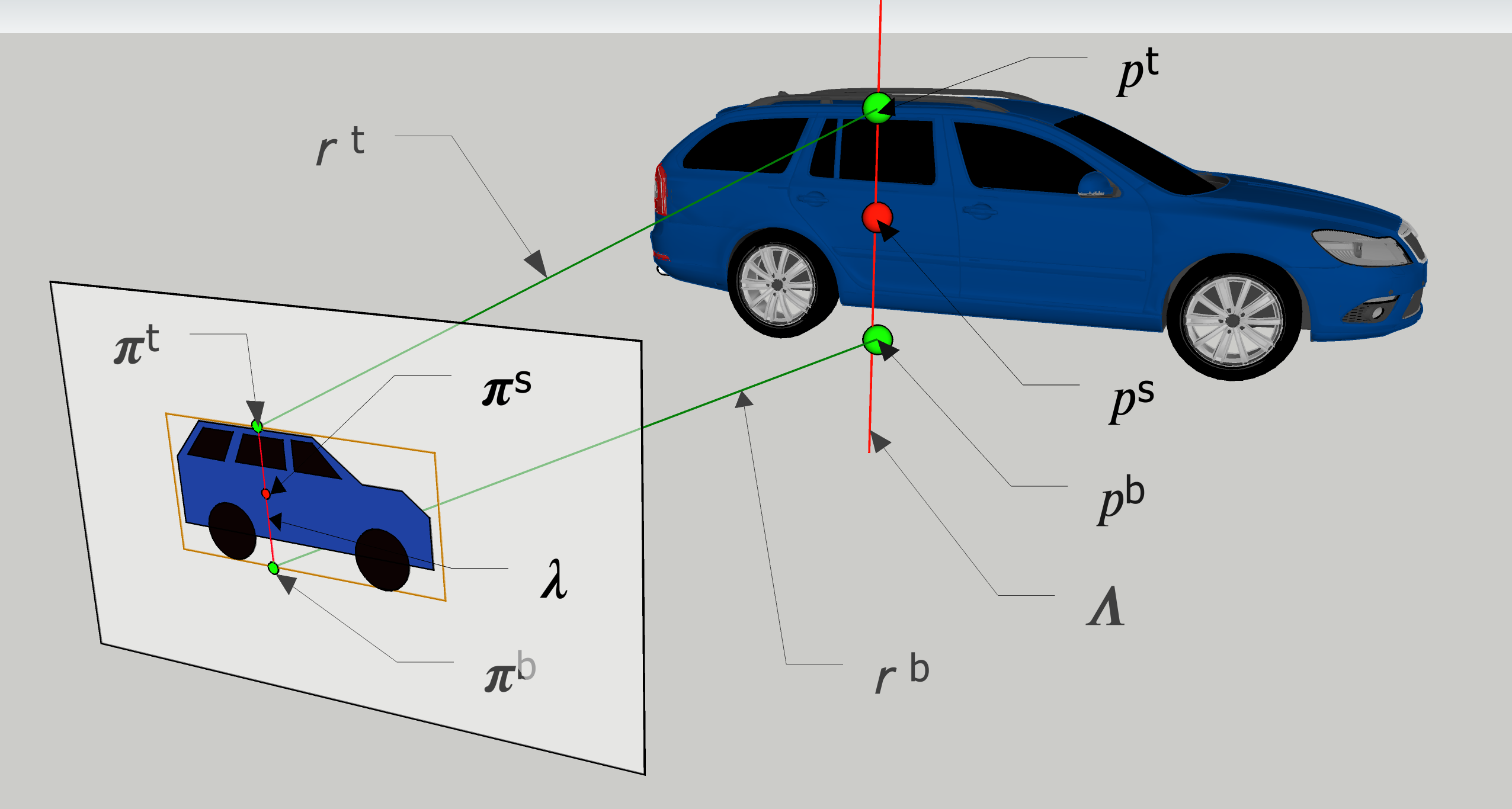}
		\end{center}
		\caption{Left: top view of a 3D object corresponding to a car detection. The dots correspond to 3D points that project inside the detection region. The green dots lie on the surface of the object, the yellow dots are closer to the camera than the object's surface (they could correspond to an occluded part of the car), and the blue dots are further away to the camera than the object's surface (they correspond to points in the background of the detection region). A representative of these points is obtained, the red dot, which is more likely to lie on the surface of the object; all the points are averaged with a gamma distribution evaluated at their depth ranking value. Right: projection of the object on the image. The red dot $p^s$ corresponds to a $3D$ point on the surface of the car, $\pi^s$ is the projection of this point on the image. $\pi^t$ and $\pi^b$ are the vertical extremities of the object on the image and $p^t$ and $p^b$ correspond to the extremities in the $3D$ world frame. $\Lambda$ is a line parallel to the vertical direction passing through $p^s$ and $\lambda$ is the projection of this line in the image. $r^t$ and $r^b$ correspond to the back projection of $\pi^t$ and $\pi^b$, respectively.}
        \label{image:car_proj}
\end{figure*}

We assume that the points $\{\hat{p}_c^1,\dots, \hat{p}_c^m \}$ are sorted in increasing order according to their distance to the camera position, given by the SLAM system. The point $p_c^s$ is obtained as a weighted average of $\{\hat{p}_c^1,\dots,\hat{p}_c^m \}$, giving higher weight to points closer to the camera except for a small portion of the closest points. This is done in order to filter out points that do not lie on the surface of the object, in particular points from the background inside the detection region, or points appearing due to partial occlusions. This can be observed on Figure~\ref{image:car_proj}, left, with the  points lying on the object surface in green, the points closer to the camera (which can lie, for example, on the ground) in yellow, and the points further away to the camera (e.g., on a building behind the car) in blue. Finally, $p_c^s$ is depicted in red.

	The averaging of the points $\{\hat{p}_c^1,\dots, \hat{p}_c^m \}$ is done with a gamma density $g$ on the index position, with parameters $\alpha = 1.5$, $\beta = 0.2$, which were determined to work well on practice. Hence, we can estimate $p_c^s$ as
    
\begin{equation}
    p_c^s = \frac{1}{\sum_{i=1}^m g(\frac{i}{m})} \sum_{i=1}^m g(\frac{i}{m}) \hat{p}_c^i.
\label{eq:wsum}
\end{equation}
    
    The 3D line $\Lambda$ is defined as the line passing through $p_c^s$ with vertical direction (see Fig.~\ref{image:car_proj}, right). Let $\pi_k^s$  be the projection of $p_c^s$ on the image with the current camera parameters and $\lambda$ a line in the image passing through $\pi_k^s$ and such that the plane obtained by back projecting $\lambda$ is vertical. 
   
    We consider the intersections of this line with the boundary of the detection $R(D^1_k)$, $\mathbf \pi^t, \mathbf \pi^b = \lambda \cap \partial R(D_k^1)$, as depicted in Figure~\ref{image:car_proj} with green dots on the image plane, while $\mathbf{\pi}^s_k$ is the red dot. These two image points are taken as the vertical extremities of the object.  
    
    Let $ r^t$ and $ r^b$ be the 3D map rays obtained by back projecting the image points $\mathbf \pi^t$ and $\mathbf \pi^b$, respectively. We define  $\tilde{p}^t = r^t \cap \Lambda$ and  $\tilde{p}^b = r^b \cap \Lambda$. These two points are taken as the vertical extremities of the object in the 3D map, as seen in Figure~\ref{image:car_proj} (in green).
    
	Then the object height can be estimated as the Euclidean distance $\hat{H} = D(\tilde{p}^t,\tilde{p}^b)$. The scale correction observation, given $H$, is calculated as $\hat{\kappa}_k = \frac{H}{\hat{H}}$. Finally, the likelihood of the detection $D_k^1$ is evaluated as 
    
    $$
  p(D_k^1|H,\kappa_k,\mathcal{N}_k) = f(\hat{\kappa}_k;\kappa_k,\sigma^m_{k})
    $$ 
    
\noindent
with $f(;m,s)$ a Gaussian density with mean $m$ and standard deviation $s$. The next section describes how $\sigma^m_{k}$ can be evaluated at $k$.
    
\subsection{Observation noise variances}

\label{subsec:sigmas}

We define $\sigma^m_{k}$, the observation noise, to quantify roughly the uncertainty on each scale observation. We evaluate it as the standard deviation on $\hat{\kappa}_k$ using uncertainty propagation, as follows. Let $d_i = \|\hat{p}_c^i\|$, with $\hat{p}_c^i$ one of the points in $\mathcal{N}_k$ as defined in the previous section, expressed in the camera frame. The variance on the depth of $p_c^s$, $\|p_c^s\|$, is approximated as the variance of the distances $d_i$ with the weights of Eq.~\ref{eq:wsum}. We refer to it as $\sigma_d^2$.

The distance $\hat{H}=D(\tilde{p}^t,\tilde{p}^b)$ can be expressed as $\|p_c^s\| |m_2 \pm m_1|$ with $m_2= \frac{D(\tilde{p}^t, p_c^s)}{\|p_c^s\|}$ and $m_1= \frac{D(\tilde{p}^b, p_c^s)}{\|p_c^s\|}$ ($m_2$ and $m_1$ are considered as constant, here, as they are the slopes of $r_t$ and $r_d$ and do not depend on $\|p_c^s\|$). Hence, the standard deviation on $D(\tilde{p}^t, p_c^s)$ is roughly $\sigma_D = |m_2\pm m_1| \sigma_d=\frac{D(\tilde{p}^t,\tilde{p}^b)}{\|p_c^s\|} \sigma_d$. Now $\hat{\kappa}_k = \frac{H}{D(\tilde{p}^t,\tilde{p}^b)}$ so $\sigma^m_{k}$ can be approximated as

\begin{equation}
\sigma^m_{k} \approx \frac{H}{\hat{H}}\frac{\sigma_d}{\|p_c^s\|}.   
\label{eq:sigmaObs}
\end{equation}

\begin{figure}[t]
	\begin{center}
    (a) \includegraphics[width=\linewidth]{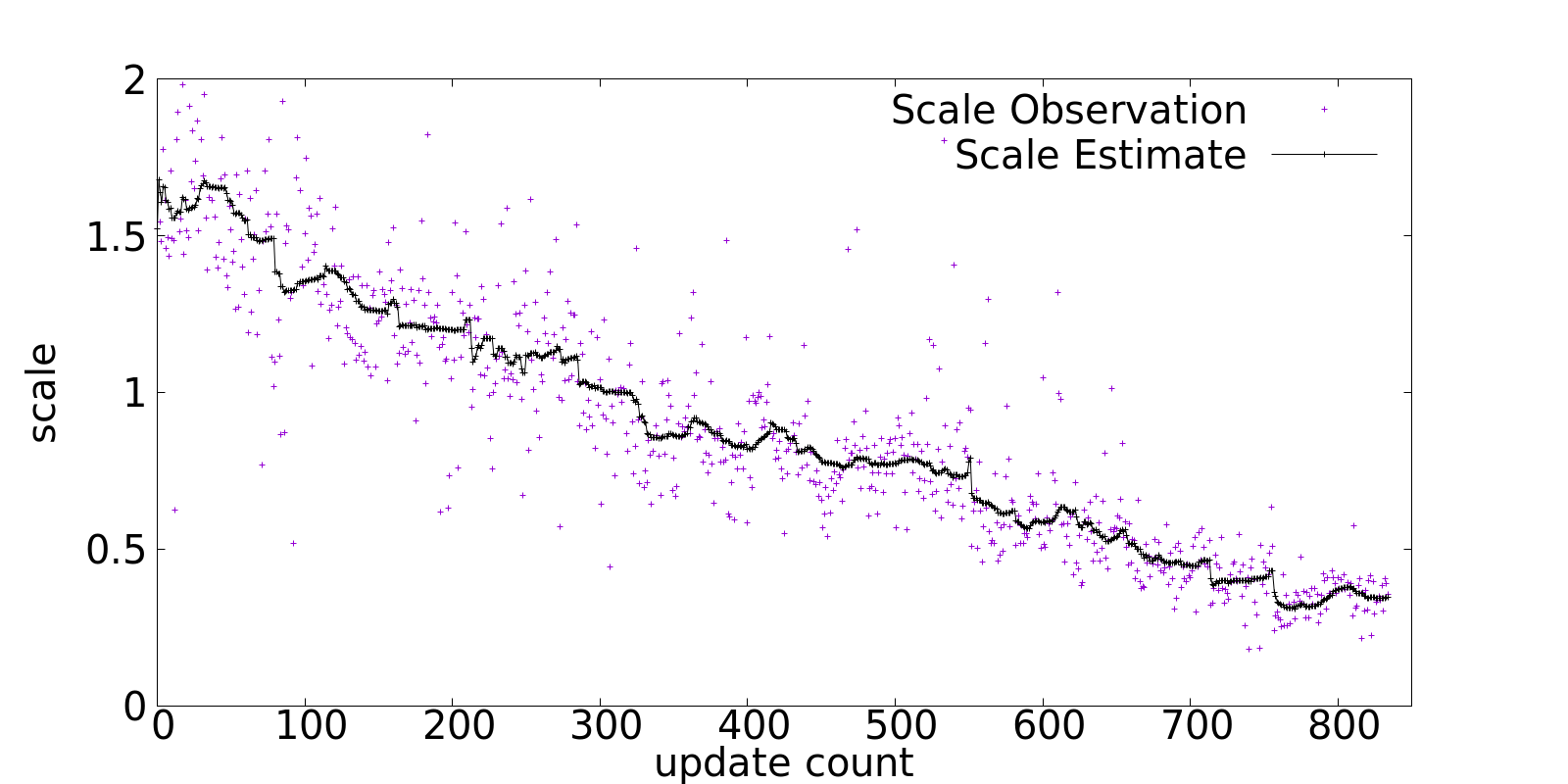}
	(b) \includegraphics[width=\linewidth]{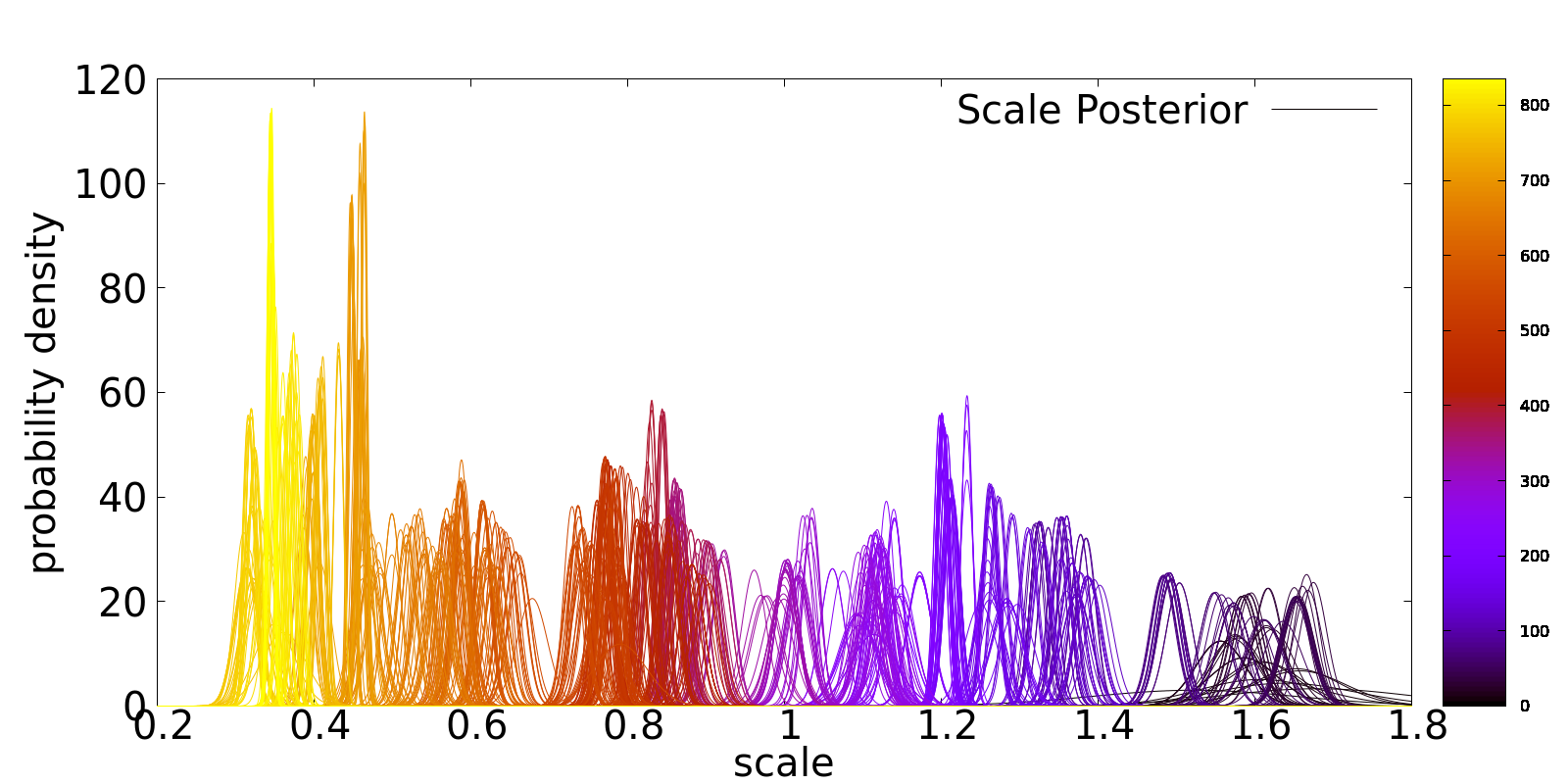}\\
	(c) \includegraphics[width=\linewidth]{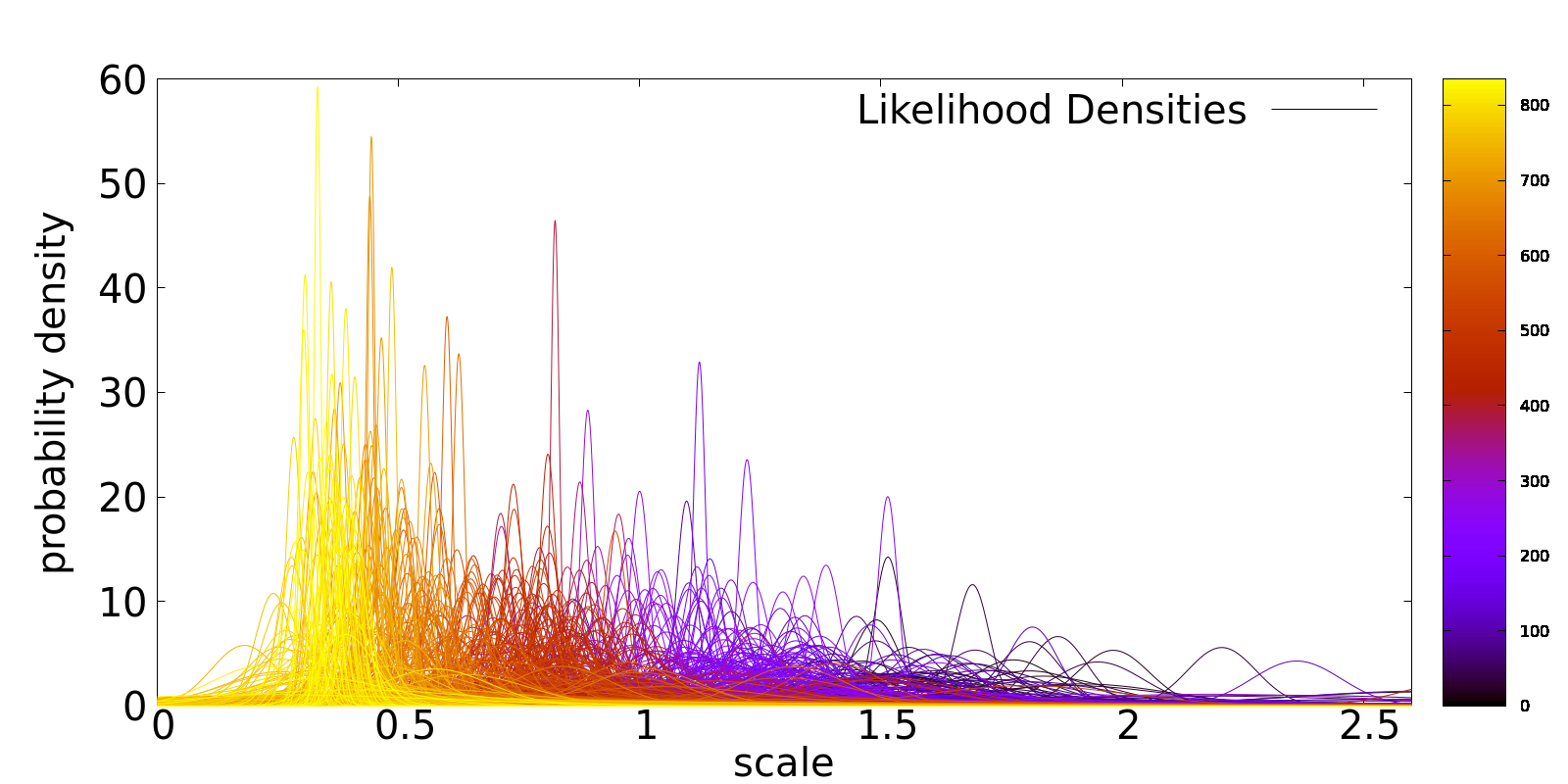}\\
		\end{center}
\caption{
(a) Evolution of the scale MAP estimate (in bold) along scales observations (the graphs correspond to KITTI sequence $00$).
(b) Evolution of the scale correction posterior. The color palette indicates the point in time (the clearer, the later in the video).
(c) Likelihoods of new observations along time. We observe likelihoods with different dispersions since the dispersion is calculated from the variance on the object's surface depth. }
\label{fig:graphs1}
\end{figure}

\subsection{Posterior updates}

\label{subsection:Kalman}
In the case the scale correction and height prior distributions are represented as discrete distributions, the implementation of Eq.~\ref{eq:update} is quite straightforward.

In the case the height prior distributions are Gaussian $f(H;\bar{H},\sigma_{c(D^1_k)}^H)$, the Bayes Filter can be implemented as a Kalman Filter, where the current scale correction estimate is represented by its mean/variance before and after correction, $\kappa_k^-,\kappa_k^+$ (means) and $\sigma^{-2}_{k},\sigma^{+2}_{k}$ (variances) with the following equations:

\subsubsection{Prediction}

It results from Eq.~\ref{eq:transmodel},  

\begin{equation}
\kappa_{k+1}^- = \kappa_k^+,
\label{eq::pred_mean}
\end{equation}
\begin{equation}
\sigma^{-2}_{k+1} = \sqrt{\sigma_k^{+2} + {\sigma_{k+1}^{p2}}}.
\label{eq::pred_sd}
\end{equation}

\subsubsection{Correction}

In this step, the difference with the traditional Kalman filter is that we have to marginalize the variable $H$ over the prior on the object class, i.e. compute $\int_{H}  p(D_k^l|H,\kappa_k,\mathcal{N}_k) p_{c(D^l_k)}(H)dH$. To simplify the evaluation and keep the result as a Gaussian, we use Eq.~\ref{eq:sigmaObs} for a fixed value of $H$, $\bar{H}$, i.e., $\sigma^m_k=\frac{\bar{H}}{\hat{H}}\frac{\sigma_d}{\|p_c^s\|}$.

In that case, we deduce that $\int_{H}  p(D_k^l|H,\kappa_k,\mathcal{N}_k) p_{c(D^l_k)}(H)dH$ is a Gaussian centered at $\frac{\bar{H}}{\hat{H}}$ with variance $\frac{1}{\hat{H}^2}(\sigma_H^2+\frac{\sigma^2_d\bar{H}^2}{\|p^s_c\|^2})$. The update expressions follow for the means and variances:

\begin{equation}
\frac{1}{\sigma^{+2}_{k+1}} = \frac{1}{\sigma^{-2}_{k+1}} + \frac{\hat{H}^2}{\sigma_H^2+\frac{\sigma^2_d\bar{H}^2}{\|p^s_c\|^2}}, 
\label{eq::update_mean}
\end{equation}
\begin{equation}
\kappa^+_{k+1} = \kappa^-_{k+1} \frac{\sigma^{+2}_{k+1}}{\sigma^{-2}_{k+1}}
+ \hat{\kappa}_k \frac{\sigma^{+2}_{k+1} \hat{H}^2}{\sigma_H^2+\frac{\sigma^2_d\bar{H}^2}{\|p^s_c\|^2}}. 
\label{eq::update_sd}
\end{equation}

\section{Experimental results}

\label{sec:experiments}

\subsection{Description of the experimental setup}

For a quantitative evaluation of the scale estimation for correcting scale drift, the algorithm is run on 10 sequences of the KITTI dataset~\cite{Geiger2012}. Each sequence consists of a driving scenario in an urban environment with varying speeds and distances. We want to stress that, although the application presented in these experiments is quite specific (monocular vision for road vehicles), the proposed method is much more generic and can be used in many other scenarios. We have chosen this application to measure its potential benefits, because of the existence of well documented datasets, such as KITTI. Sequences $00$ to $10$ are considered here, except for sequence $01$, since it is in a highway in which the SLAM algorithm (ORB-SLAM) fails due to the high speed. The sequences come with ground truth poses for evaluation. The evaluation computes errors between relative transformations for every subsequence of length $(100,200,...,800)$ meters as proposed in~\cite{Geiger2012}. Here, as our algorithm evaluates the scale correction, we only present results on translational errors. The rotational errors are a consequence of the SLAM algorithm and do not depend on the scale. 

\subsection{Implementation}

The monocular version of ORB-SLAM 2 \cite{murORB2} is used for tracking and mapping. Loop closure is disabled so that the scale drift is directly observed (as in Fig.~\ref{fig:summary}). 

YOLO9000~\cite{yolo} is used for detecting car instances, and the minimum confidence threshold is set to $0.45$. We could have considered more object classes but their presence in the KITTI dataset is marginal (a few ``truck'' or ``bus'' objects only). Object detection is run every 5 frames. As it can be seen in Table~\ref{table::stats}, the number of updates, i.e. of integrations of observations in the Bayesian framework, is quite variable. In sequence $00$ or $07$, there are approximately $0.2$ updates per frame; in sequence $04$, this number falls to $0.014$. Of course, this has an impact on the final errors (see below).

The prior distribution for the car's height is set as a Gaussian with mean $1.5$ meters. The mean was chosen in accordance with the report by the International Council on Clean Transportation~\cite{eur15} for average car height in 2015. Based on these facts, we selected the Kalman Filter implementation of the algorithm, equations \ref{eq::pred_mean} and \ref{eq::pred_sd} for prediction and equations \ref{eq::update_mean} and \ref{eq::update_sd} for correction.  

The ORB-SLAM and YOLO algorithms run in real time, and the Kalman filter implementation of the scale correction estimation adds negligible additional processing time, which guarantees the real time performance of the algorithm.

\label{subsection:descriptionexp}

\subsection{Evaluation and discussion}
{\small
\begin{table}[t]
\centering
\begin{tabular}{|l|l|l|l|}
\hline
Sequence & Frames & Updates & Avg. updates   \\ 
 &  &  &  per frame   \\\hline
00 & 4540 & 921 & 0.202 \\ \hline
02 & 4660 & 289 & 0.062  \\ \hline
03 & 800 & 24 & 0.03   \\ \hline
04 & 270 & 4 & 0.014 \\ \hline
05 & 2760 & 202 & 0.073  \\ \hline
06 & 1100 & 75 & 0.068  \\ \hline
07 & 1100 & 234 & 0.212 \\ \hline
08 & 4070 & 530 & 0.130 \\ \hline
09 & 1590 & 111 & 0.069  \\ \hline
10 & 1200 & 40 & 0.033  \\ \hline
\end{tabular}
\caption{Scale update statistics for 10 sequences of the KITTI dataset.\label{table::stats}}
{\scriptsize
\begin{tabular}{|l|l|l|l|l|l|}
\hline
Seq. & {\tiny ORB-SLAM} & {\tiny ORB-SLAM}  & {\tiny ORB-SLAM} & {\tiny ORB-SLAM} & \cite{Song13}  \\ 
 &Bayesian & Update only & Avg. Scale & Stereo~\cite{murORB2} &  \\ 
  &(ours) &  &  & &   \\ 
  &Trans(\%) & Trans(\%) & Trans(\%) & Trans(\%)&   Trans(\%) \\ 
\hline
00 & 3.09 & 20.77 & 35.78 & 0.70 &  7.14\\ \hline
02 & 6.18 & 4.81 & 8.01 & 0.76 & 4.34 \\ \hline
03 & 3.39 & 2.82 & 15.07 & 0.71 &  2.90\\ \hline
04 & 32.90 & 26.14 & 26.19 & 0.48 &  2.45\\ \hline
05 & 4.47 & 17.54 & 51.69 & 0.40 &  8.13\\ \hline
06 & 12.46 & 16.22 & 20.883 & 0.51 &  7.56\\ \hline
07 & 2.81 & 11.66 & 10.76 & 0.50 &  9.92\\ \hline
08 & 4.11 & 20.32 & 31.88 & 1.05 &  7.29\\ \hline
09 & 11.24 & 12.69 & 22.65 & 0.87 &  5.14\\ \hline
10 & 16.75 & 16.11 & 18.65 & 0.60 &  4.99\\ \hline
Avg. & 5.52  & 15.30 & 27.89 & 0.0893 & 6.42 \\ \hline
\end{tabular}}
\caption{Comparison of relative translational error.\label{table::error}}
\end{table}
}

\begin{figure*}[t]
	\begin{center}
    \includegraphics[width=0.5\linewidth]{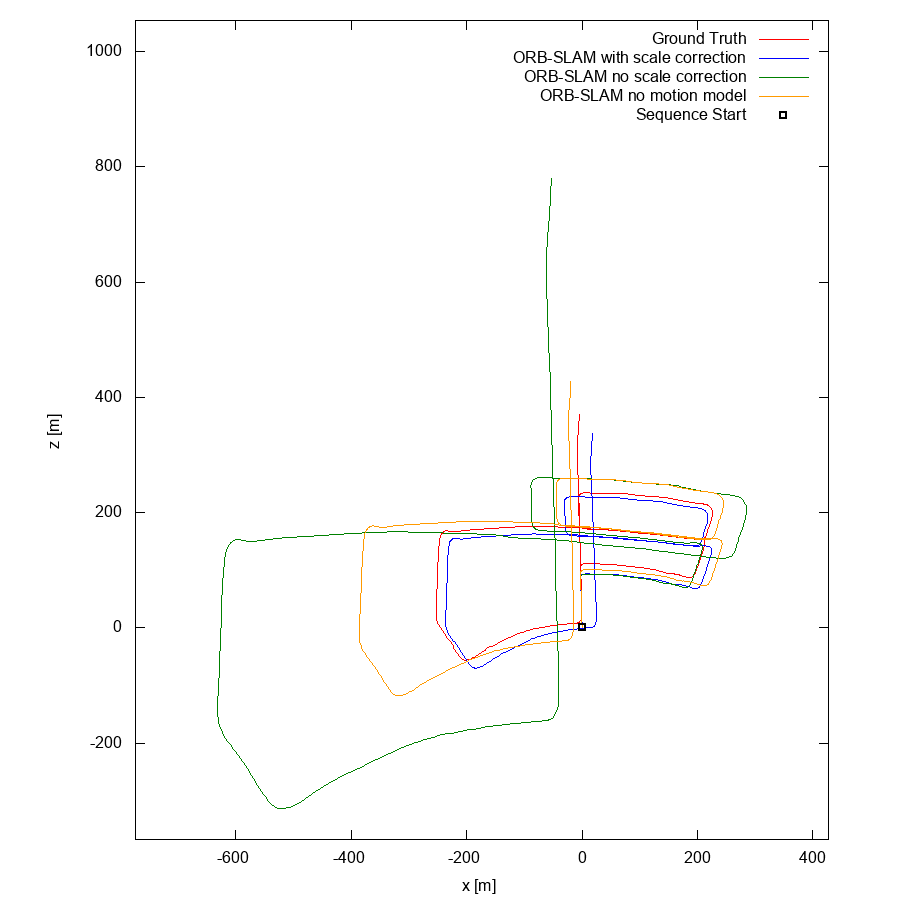}\includegraphics[width=0.5\linewidth]{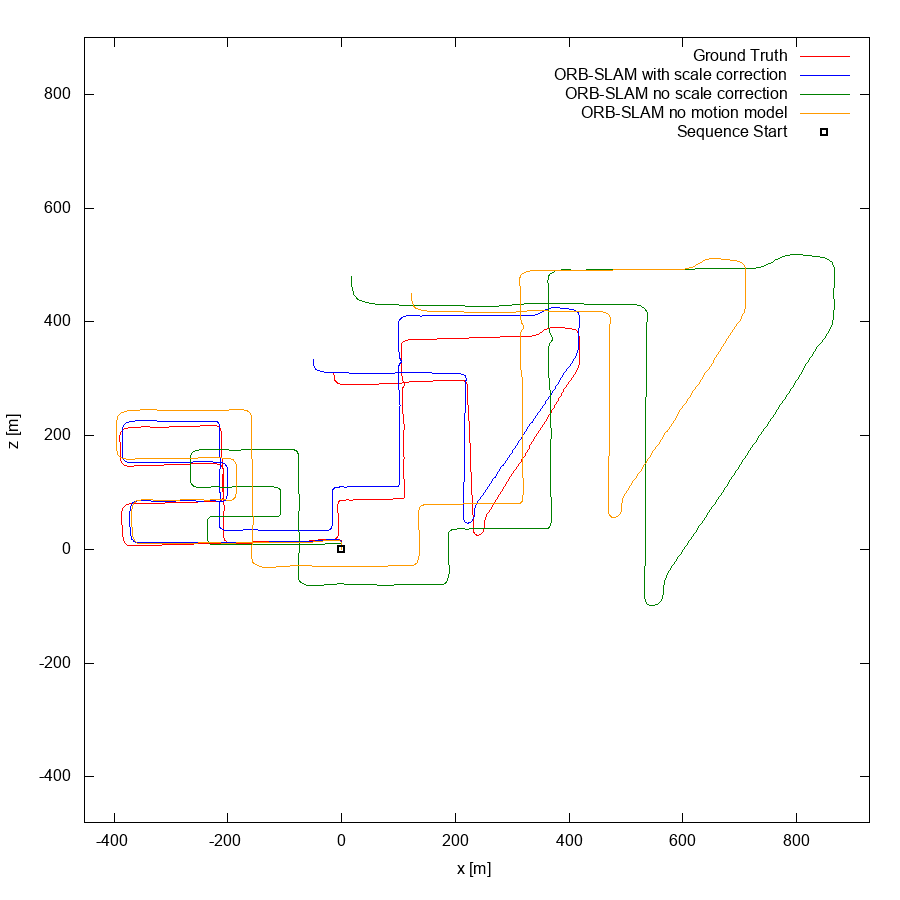}
		\end{center}
        \caption{Reconstructed trajectories for sequences $05$ (left) and $08$ (right). The experiments compare the ground truth (in red), the output of ORB-SLAM (in green), our scale correction algorithm with motion model (in blue) and without it (in orange).\label{fig:trajectories}}
\end{figure*}

\begin{figure}[t]
	\begin{center}
    \includegraphics[width=\linewidth]{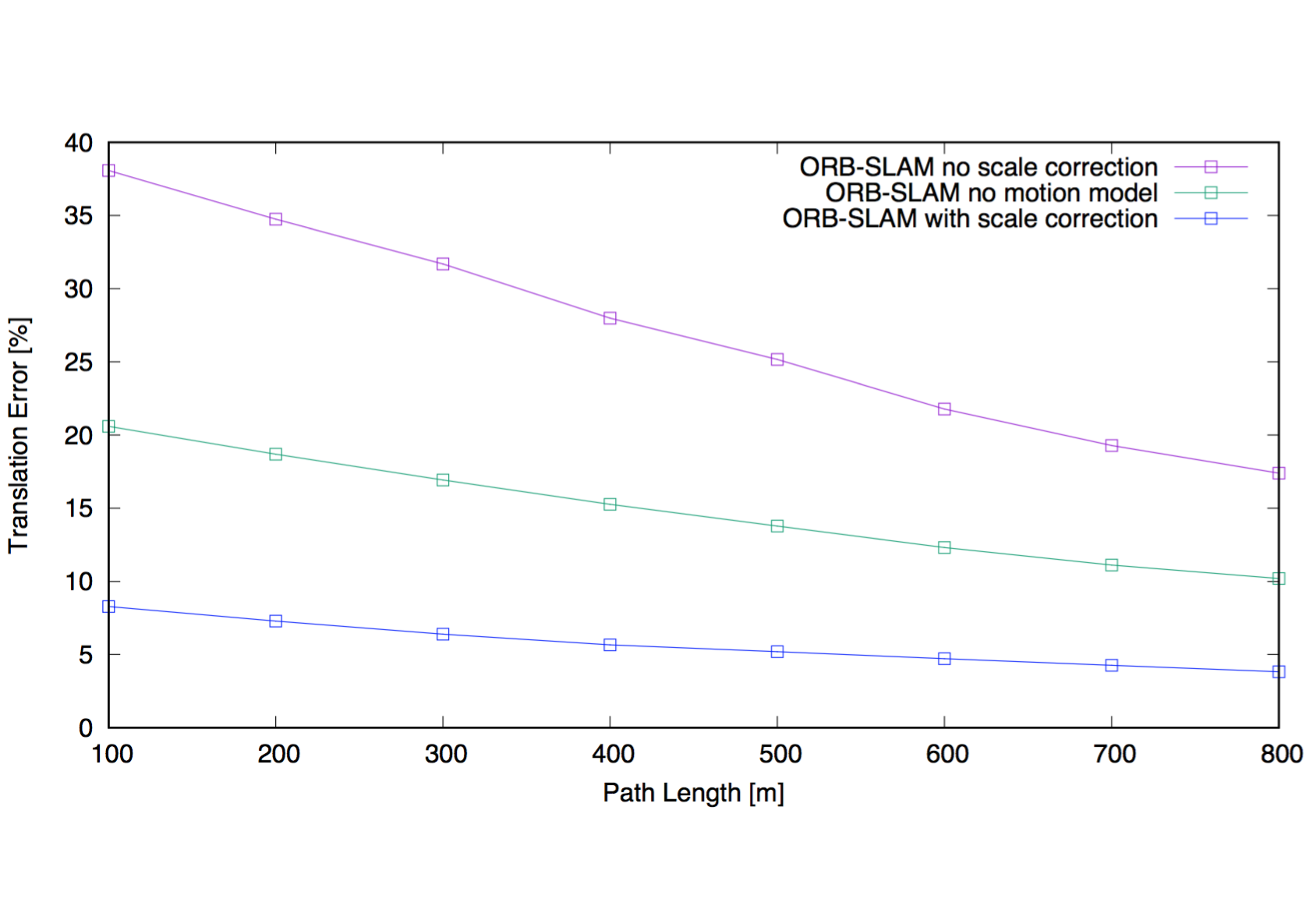}
		\end{center}
        \caption{Evolution of the average translational errors with/without scale correction, and with/without motion model for the scale correction factor, evaluated on the 10 KITTI sequences.\label{fig:avgerrors}}
\end{figure}
We can see in Figure~\ref{fig:graphs1}(a) the evolution of the scale estimate (in bold) along with the scale observations corresponding to the KITTI sequence $00$, i.e. the same as the illustration of Fig.~\ref
{fig:summary}. Our scale correction estimate is clearly decreasing from values $>1$, i.e. as ORB-SLAM is sub-estimating the scale, to values $<1$ towards the end of the sequence, i.e. as ORB-SLAM is over-estimating the scale. This effect is perceptible in Fig.~\ref
{fig:summary}, through the path estimated by ORB-SLAM: distances in the trajectory produced by ORB-SLAM without scale estimation (in green) are seen to be overestimated later in time.

Figure \ref{fig:graphs1}(b) shows the evolution in time of the scale posterior. The peaks correspond to updates with a low uncertainty, as can be seen in the plot of all update distributions in Figure~\ref{fig:graphs1}(c). The time is indicated by the color of the posterior (lighter colors means later times). Updates can also have a higher effect on the posterior in moments where a large scale drift is expected due to high rotational translation, as suggested by equation~\ref{eq::omega}.

In Table~\ref{table::error}, we compare the errors obtained with different approaches for scale estimation for the 10 KITTI sequences analyzed: (i) in the second column, our method as presented in this paper; (ii) in the third column, our method without the scale correction motion model, i.e. roughly as in~\cite{Sucar2017}; (iii) in the fourth column, a very simple method that computes an average value of the scale correction, $\frac{1}{N} \sum_{i=1}^N \hat{\kappa}_k$, and applies it to the map and the trajectory (this corresponds to neglecting the scale drift effect); (iv) in the fifth column, to give a hint on the precision reached by a 3D sensor, we give the results by ORB-SLAM 2 with the stereo datasets; finally, (v) the fifth column gives results from the monocular system developed in~\cite{Song13}, where the camera height over the road is known. 

Analyzing the results for our methods (second and third columns) in the different sequences, one can see a strong correlation between the obtained errors and the average number of updates per frame as described in Table~\ref{table::stats}, as expected. For example, in sequences $00$, $07$, $08$, where a lot of cars where detected, the results are very good, with errors significantly lower than~\cite{Song13}. On the opposite, sequences $04$, $09$, $10$ with their scarce car detections, give rather poor results. Sequence $04$, for instance, is a short sequence in a highway, without static vehicles, and produces only 4 update steps. However, (bottom row), the overall error levels are lower than~\cite{Song13}. Note that introducing the motion model with varying variance has allowed to improve the performance of~\cite{Sucar2017} by a factor of 3. Last, as expected, not including the scale drift (fourth column) leads to very poor performance. Finally, a detector such as~\cite{yolo} is quite versatile, so we could use it at its maximum potential by integrating other classes to detect, e.g. road signs, house doors and windows.       

In Fig.~\ref{fig:trajectories}, we give two more examples of reconstructed trajectories with/without our scale correction  and with/without motion model for the scale correction factor. Our method allows the final trajectory (in blue) to get very close to the ground truth (in red). Similarly, in Fig.~\ref{fig:avgerrors}, we give the errors of these same methodologies, for different path lengths, and averaged over the 10 sequences. Again, our method allows to get very reasonable errors, between 4$\%$ and 7$\%$. 

Some of the best monocular systems with scale correction, \cite{7225730} and \cite{Zhou2016ReliableSE}, have average errors of $5\%$ and $3\%$, respectively, which are very similar to the average error of our method, $5.53\%$. But these monocular methods are specific to driving scenarios, based on a given fixed camera height and an observable plane. On the other hand, state of the art methods for scale estimation based on object detection, \cite{Frost2016}, have errors of $20\%$ in average. Our method outperforms state of the art methods of scale estimation based on object detection while achieving similar performance to state of the art monocular systems with scale correction, but within a more general framework.

\section{Conclusions}

We have presented a Bayes filter algorithm that allows to estimate the scale correction to apply to the output of a monocular SLAM algorithm so as to obtain correct maps and trajectories. The observation model uses object detections given by a generic object detector, and integrates height priors over the object from the detected classes. A probabilistic motion model is proposed in order to model the scale drift. In the light of the very promising results obtained in the KITTI dataset, we will put our efforts in obtaining a better model for the scale drift, whose evolution over time seems to exhibit a clear structure.     

\bibliographystyle{ieee}
\bibliography{egbib}

\end{document}